\documentclass[11pt]{article}

\usepackage[margin=1in]{geometry}
\usepackage{graphicx}
\usepackage{booktabs}
\usepackage{array}
\usepackage{hyperref}
\usepackage{enumitem}
\usepackage{listings}
\usepackage{xcolor}
\usepackage{caption}

\graphicspath{{figures/}}
\hypersetup{
  colorlinks=true,
  linkcolor=blue,
  urlcolor=blue,
  citecolor=blue
}

\lstdefinestyle{cmd}{
  basicstyle=\ttfamily\small,
  breaklines=true,
  frame=single,
  backgroundcolor=\color{gray!6},
  columns=fullflexible
}
\title{NeuraDock Visual Cognitive Load Agent Tutorial: A Quality-Gated Open-Source EEG Workflow for Alpha Dynamics and Real-Time Applications}
\author{Zhiyuan Xu, Yueqing Dai, Junling Li and Junwen Luo\thanks{Corresponding author: Junwen Luo (technology@neuradock.com)} \\
\textit{Shanghai Pulse Element Intelligent Technology Co., Ltd. (NeuraDock)}}
\date{Tutorial manuscript, updated for release 2026.06.24}

\begin{document}

\maketitle

\begin{center}
\textbf{Open-source agent and latest instructions:}\\
\url{https://github.com/Neuradock/eeg-workstation-agent}
\end{center}

\begin{abstract}
This tutorial paper provides a step-by-step, reproducible walkthrough of NeuraDock Agent, an open-source EEG agent focused on Alpha dynamics and visual cognitive-load analysis. The goal is practical: a reader should be able to install the agent, run EEG preprocessing and quality control, generate Alpha dynamics figures, perform within-subject Rest/Task visual cognitive-load comparison, run the public mini-dataset analyses and compare them with the reference validation summary, start an online dashboard, call the real-time API from an external application, and use the LLM interpretation layer to explain quality risks. Existing EEG toolkits provide excellent offline analysis, but assembling a real-time, quality-gated cognitive-load pipeline often requires manually bridging acquisition, custom QC, Alpha feature extraction, and a web API; this tutorial closes that offline-to-online gap. The tutorial uses a quality-gated workflow: downstream Alpha and workload metrics are computed only after preprocessing and QC gating rather than directly from raw EEG. In the included mini-dataset validation, the agent processed 18 recordings, generated 10 within-subject comparisons, observed task-related posterior Alpha suppression in 7 of 10 contrasts, estimated initial evidence of within-subject repeatability, and benchmarked local online API latency. The tutorial is intended for researchers, developers, and applied teams who want a transparent path from EEG files to real-time visual cognitive-load prototypes.
\end{abstract}

\noindent\textbf{Keywords:} EEG tutorial; visual cognitive load; Alpha dynamics; posterior Alpha suppression; open source; real-time API; quality control; NeuraDock Agent

\section{Reader Goal}

This tutorial is written as a hands-on paper. By the end, a reader should be able to reproduce the following outputs:

\begin{itemize}[leftmargin=*]
  \item EEG preprocessing and signal-quality reports.
  \item Clean/QC-gated EEG output files.
  \item Alpha dynamics time-domain, frequency-domain, and time-frequency figures.
  \item Within-subject offline Rest/Task cognitive-load comparison figures.
  \item Mini-dataset per-recording and within-subject comparison outputs.
  \item A real-time local dashboard and API endpoint.
  \item LLM-based explanations of computed EEG results and quality risks.
\end{itemize}

More importantly, by the end of the tutorial the reader will have a local HTTP endpoint that streams a visual workload index and quality flag to an application. Wiring together that path from offline EEG scripts to a live API is often the part that turns a promising cognitive-load idea into weeks of infrastructure work.

The tutorial also states what the workflow is not: it is not a clinical diagnostic system, not an attention or fatigue diagnosis, and not a cross-subject cognitive-ability ranking method.

\section{Why a Tutorial Workflow Is Needed}

General EEG tools such as MNE-Python, EEGLAB, and BrainFlow are powerful and open source, but they are broad toolkits rather than focused cognitive-load workflows \cite{mnepython,gramfort2013,eeglab,delorme2004,brainflow}. Commercial and device-specific systems such as Emotiv Cortex and Neurosity SDK workflows provide real-time data access, but they are not fully open, locally auditable cognitive-load analysis stacks \cite{emotiv,neurosity}. Existing cognitive-load studies often remain offline and do not provide a complete path from preprocessing to a real-time API. This creates the practical ``why now'' gap: teams building adaptive interfaces, XR demos, HMI experiments, or internal workload dashboards can analyze EEG offline, but still struggle to deploy the same logic as a quality-gated service that another application can call once per second.

NeuraDock Agent fills this gap by packaging a focused tutorial workflow:

\begin{lstlisting}
Install
  -> Check profile
  -> Preprocess and quality-gate EEG
  -> Analyze Alpha dynamics
  -> Compare Rest/Task within subject
  -> Run public mini-dataset analyses
  -> Start online dashboard/API
  -> Interpret results with LLM summary layer
\end{lstlisting}

The value is not only a feature extractor. The value is an auditable, repeatable route from EEG data to cognitive-load application prototypes.

\section{Tutorial Assumptions}

The commands below assume Windows PowerShell and that the current directory is
the root of the cloned \texttt{eeg-workstation-agent} repository. All paths are
relative to that repository; no machine-specific installation path is
required.

The short examples use public files downloaded and SHA256-verified by the
repository's data helper:

\begin{lstlisting}
data_examples\alpha\open_closed_eye2.txt
data_examples\rest_task\rest_S01_1.txt
data_examples\rest_task\task_S01_1.txt
\end{lstlisting}

The complete tutorial dataset is maintained separately at
\url{https://github.com/Neuradock/eeg-workstation-data/tree/add-visual-cognitive-load-mini-dataset-20260622/visual_cognitive_load/mini_dataset_v20260622}.

The workflow uses a seven-channel NeuraDock profile and a 250 Hz sampling rate. The posterior and parieto-occipital channels are used for visual Alpha features.

\section{Step 1: Install and Check the Agent}

Clone the open-source repository and enter its root directory:

\begin{lstlisting}[style=cmd]
git clone https://github.com/Neuradock/eeg-workstation-agent.git
cd eeg-workstation-agent
\end{lstlisting}

Create a virtual environment and install the agent:

\begin{lstlisting}[style=cmd]
py -m venv .venv
.\.venv\Scripts\python.exe -m pip install --upgrade pip
.\.venv\Scripts\python.exe -m pip install -e ".[dev]"
\end{lstlisting}

Download the three public example recordings used in Steps 2--4:

\begin{lstlisting}[style=cmd]
.\.venv\Scripts\python.exe scripts\download_example_data.py
\end{lstlisting}

The downloader verifies each file against a release-pinned SHA256 value.
Human EEG recordings are maintained in the separate data repository; the MIT
software license does not automatically replace the data repository's usage
and redistribution terms.

Check the version and hardware profile:

\begin{lstlisting}[style=cmd]
.\.venv\Scripts\neuradock-agent.exe --version
.\.venv\Scripts\neuradock-agent.exe profile
\end{lstlisting}

Expected version:

\begin{lstlisting}
2026.6.24
\end{lstlisting}

The actual hardware profile output from this tutorial run was:

\begin{lstlisting}
{
  "profile_version": "1.1",
  "device": "NeuraDock EEG Workstation",
  "hardware_revision": "public-profile-v2",
  "sampling_rate_hz": 250,
  "channels": ["CP5", "CP6", "PO3", "PO4", "O1", "Oz", "O2"],
  "amplitude_unit": "microvolts",
  "line_frequency_hz": 50.0,
  "tcp_default_ip": "127.0.0.1",
  "tcp_default_port": 9600,
  "tcp_start_command": "start",
  "packet_total_channels": 8,
  "packet_used_channels": 7,
  "bluetooth_samples_per_packet": 5,
  "quality": {
    "line_noise_power": 10.0,
    "emg_power": 20.0,
    "outlier_count_per_second": 2,
    "outlier_absolute_amplitude": 100.0,
    "bad_channel_segment_ratio": 0.4,
    "minimum_neighbor_correlation": 0.15
  },
  "channel_count": 7
}
\end{lstlisting}

This confirms that the tutorial uses a 7-channel profile sampled at 250 Hz, with posterior channels \texttt{PO3}, \texttt{PO4}, \texttt{O1}, \texttt{Oz}, and \texttt{O2} available for visual Alpha features. The TCP defaults also explain why online mode can start from only an IP address and port.

\subsection{Hardware-Tuned Design}

The NeuraDock montage is not arbitrary for this tutorial. The posterior ring \texttt{PO3/PO4/O1/Oz/O2} targets occipital and parieto-occipital Alpha dynamics, which are the primary signal family used here for visual cognitive-load estimation. The lateral channels \texttt{CP5/CP6} provide additional spatial context for quality checks and posterior asymmetry interpretation. The agent can ingest generic 7-channel data with the same shape, but the spatial QC thresholds, posterior-channel grouping, and right-minus-left asymmetry metrics are calibrated for this NeuraDock geometry. For best results, use the NeuraDock hardware profile rather than treating the agent as an arbitrary EEG file parser.

\section{Step 2: Run EEG Preprocessing and Quality Control}

Preprocessing is the first scientific checkpoint. Run:

\begin{lstlisting}[style=cmd]
.\.venv\Scripts\neuradock-agent.exe analyze `
  data_examples\alpha\open_closed_eye2.txt `
  --workflow quality
\end{lstlisting}

Typical output structure:

\begin{lstlisting}
runs/<timestamp>_quality/
|-- report.md
|-- results.json
|-- clean_eeg_data.npz
`-- figures/
    |-- signal_quality.png
    `-- clean_signal.png
\end{lstlisting}

How to read this step:

\begin{itemize}[leftmargin=*]
  \item \texttt{report.md} is the human-readable QC report.
  \item \texttt{results.json} contains retention rate, warning messages, rejected segment information, and bad-channel candidates.
  \item \texttt{clean\_eeg\_data.npz} stores the clean/QC-gated EEG data.
  \item The figures help inspect retained signal quality and rejected portions.
\end{itemize}

The actual tutorial run on \texttt{open\_closed\_eye2.txt} produced:

\begin{lstlisting}
Summary : signal_quality: retained 65.9%, status=warning
raw_shape              : [7, 13210]
clean_shape            : [7, 8710]
retention_rate         : 0.659
rejected_segment_count : 18
bad_channel_candidates : []
warnings:
  - Only 65.9% of samples passed segment QC.
  - Body activity heuristic: slight_activity_or_mild_muscle_tension.
\end{lstlisting}

The result is usable for tutorial purposes, but it is not a perfectly clean recording. The agent explicitly marks the run as \texttt{warning} because roughly one third of samples were rejected by segment-level QC. This is the desired behavior: downstream Alpha analysis should inherit this caution rather than pretending the signal was clean.

\begin{figure}[htbp]
  \centering
  \includegraphics[width=\linewidth]{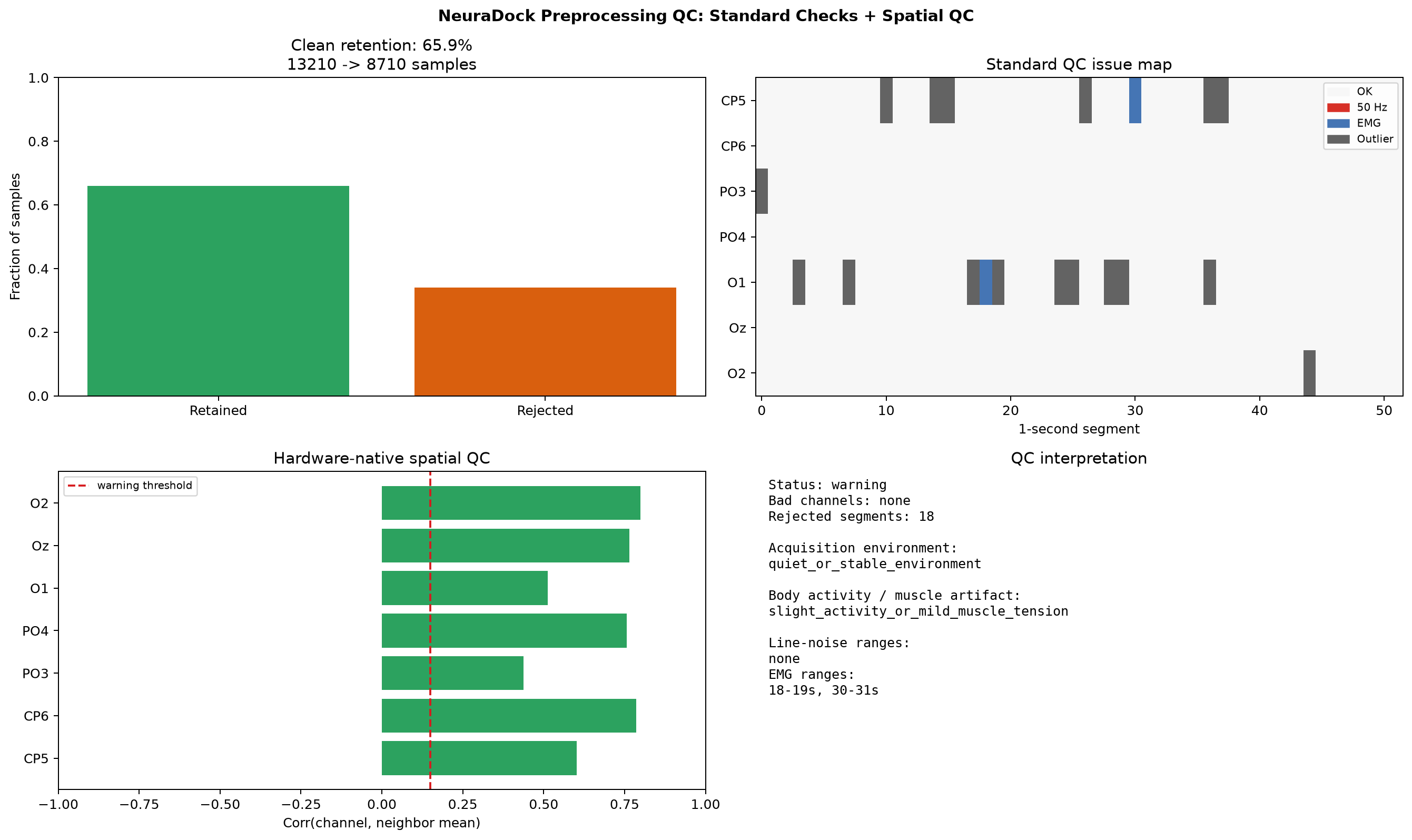}
  \caption{Signal-quality figure from the actual \texttt{open\_closed\_eye2.txt} preprocessing run. The report retained 65.9\% of samples and flagged mild body activity or muscle tension.}
  \label{fig:tutorial-signal-quality}
\end{figure}

\begin{figure}[htbp]
  \centering
  \includegraphics[width=\linewidth]{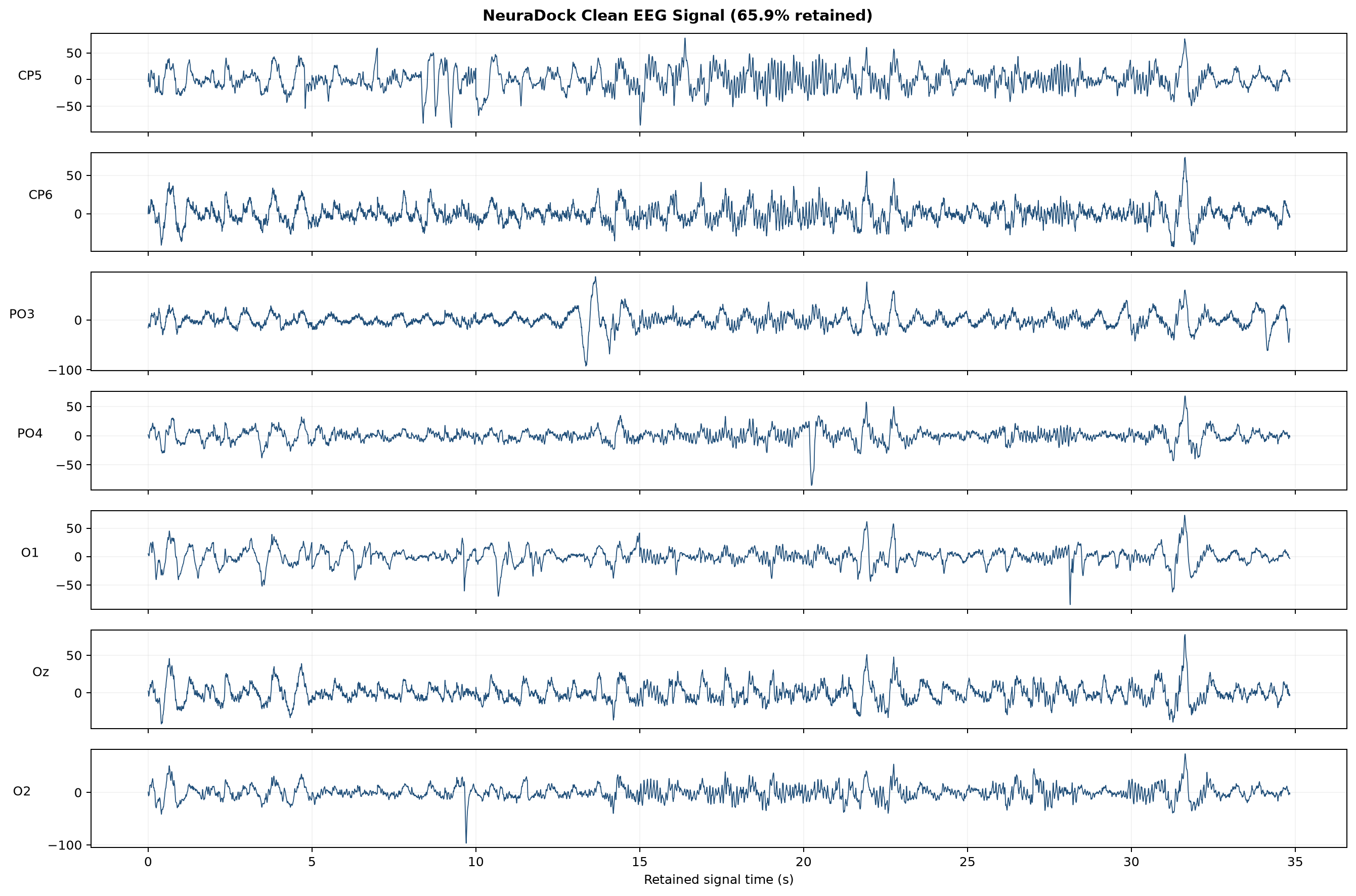}
  \caption{Clean EEG signal after preprocessing and QC gating for \texttt{open\_closed\_eye2.txt}. Later tutorial metrics are based on this quality-gated workflow rather than raw EEG alone.}
  \label{fig:tutorial-clean-signal}
\end{figure}

This is a core design principle: later Alpha and workload metrics are based on the preprocessed/QC-gated workflow, not directly on raw EEG.

\section{Step 3: Analyze Alpha Dynamics}

Run the Alpha dynamics workflow:

\begin{lstlisting}[style=cmd]
.\.venv\Scripts\neuradock-agent.exe analyze `
  data_examples\alpha\open_closed_eye2.txt `
  --workflow alpha dynamics
\end{lstlisting}

Typical output structure:

\begin{lstlisting}
runs/<timestamp>_alpha_dynamics/
|-- report.md
|-- results.json
`-- figures/
    |-- alpha_time_domain.png
    |-- alpha_frequency_domain.png
    `-- alpha_time_frequency.png
\end{lstlisting}

The Alpha dynamics workflow automatically identifies weak Alpha, baseline Alpha, strong Alpha, Alpha suppression from baseline, peak Alpha frequency, and right-minus-left Alpha asymmetry. The actual \texttt{open\_closed\_eye2.txt} run produced:

\begin{lstlisting}
Summary : alpha_dynamics: weak=4, strong=4, max suppression=+0.731
quality_status                         : warning
retention_rate                         : 0.659
valid_window_count                     : 10
excluded_window_count                  : 39
weak_alpha / baseline_alpha / strong   : 4 / 2 / 4
max_alpha_suppression_from_baseline    : 0.731
median_alpha_peak_hz                   : 10.25
median_alpha_asymmetry_right_minus_left: 0.309
warnings:
  - Only 65.9% of samples passed segment QC.
  - 39 Alpha windows were excluded because clean retention was below 80%.
\end{lstlisting}

This output is intentionally conservative. Only 10 Alpha windows were quality-valid, so the Alpha dynamics are useful as a demonstration of the feature pipeline but should be interpreted with the QC warning in mind.

\begin{figure}[htbp]
  \centering
  \includegraphics[width=\linewidth]{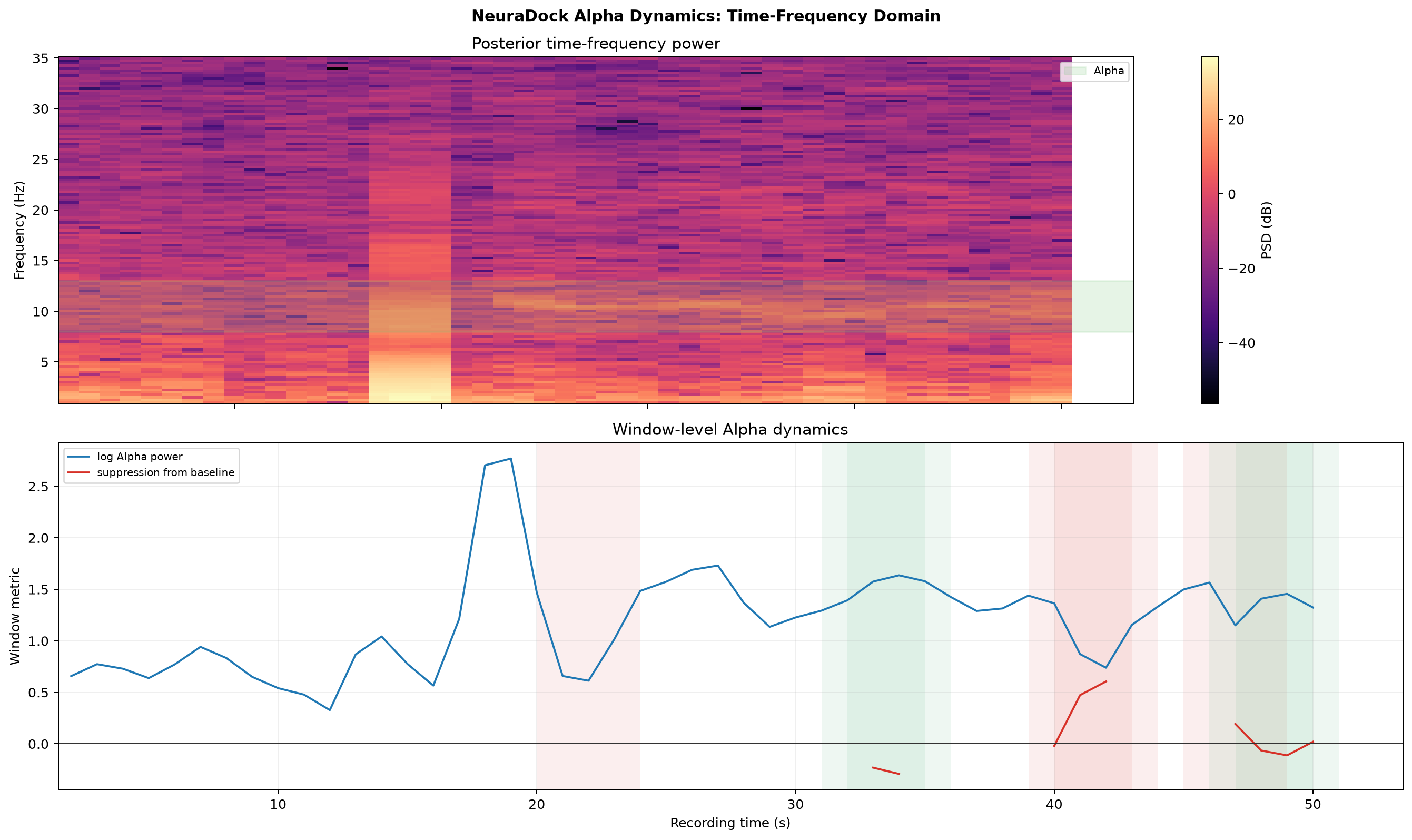}
  \caption{Actual \texttt{open\_closed\_eye2.txt} Alpha dynamics time-frequency output. This replaces the earlier task-recording example and matches the command in this tutorial step.}
  \label{fig:alpha-time-frequency-tutorial}
\end{figure}

\begin{figure}[htbp]
  \centering
  \includegraphics[width=0.49\linewidth]{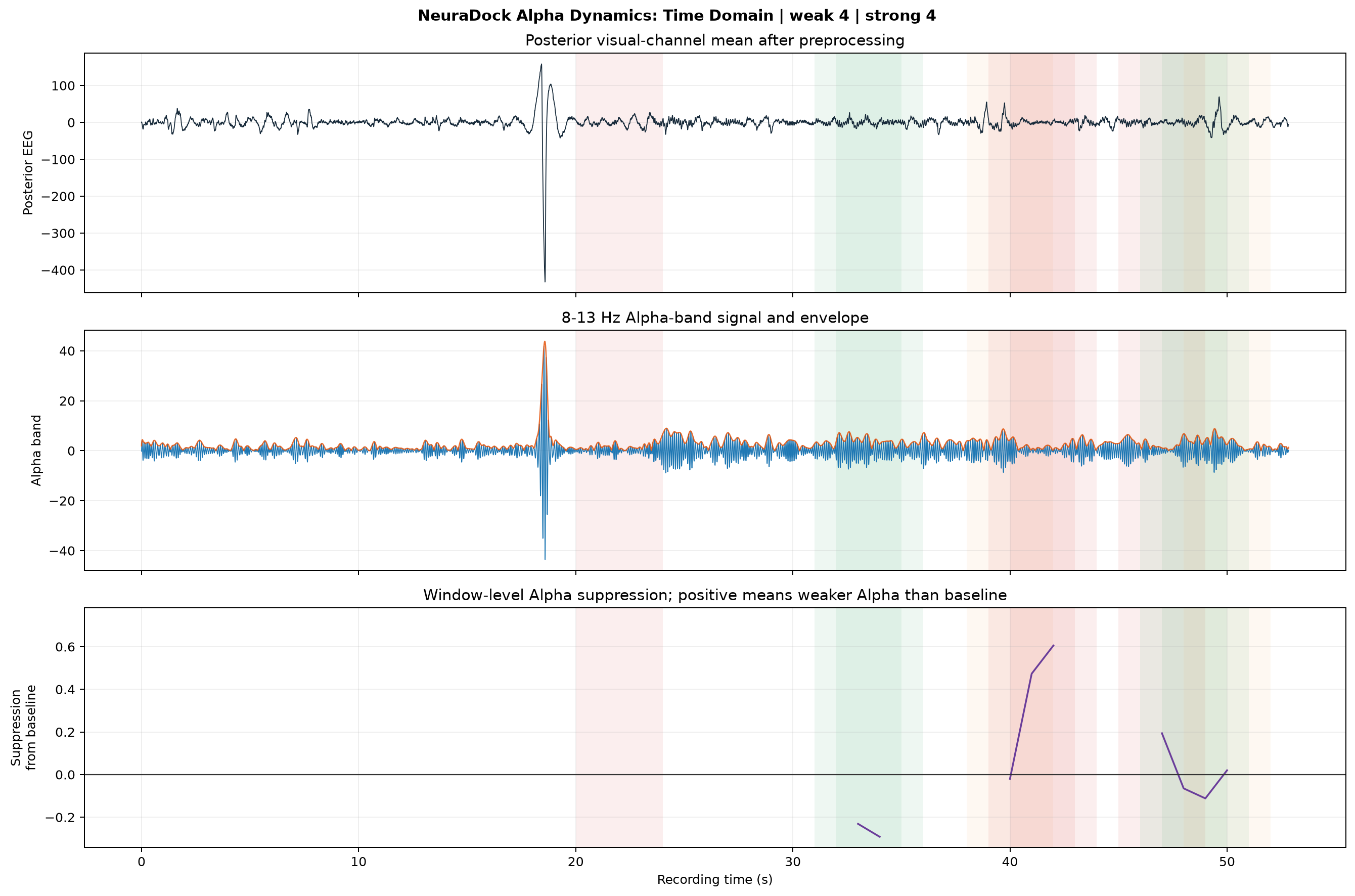}
  \includegraphics[width=0.49\linewidth]{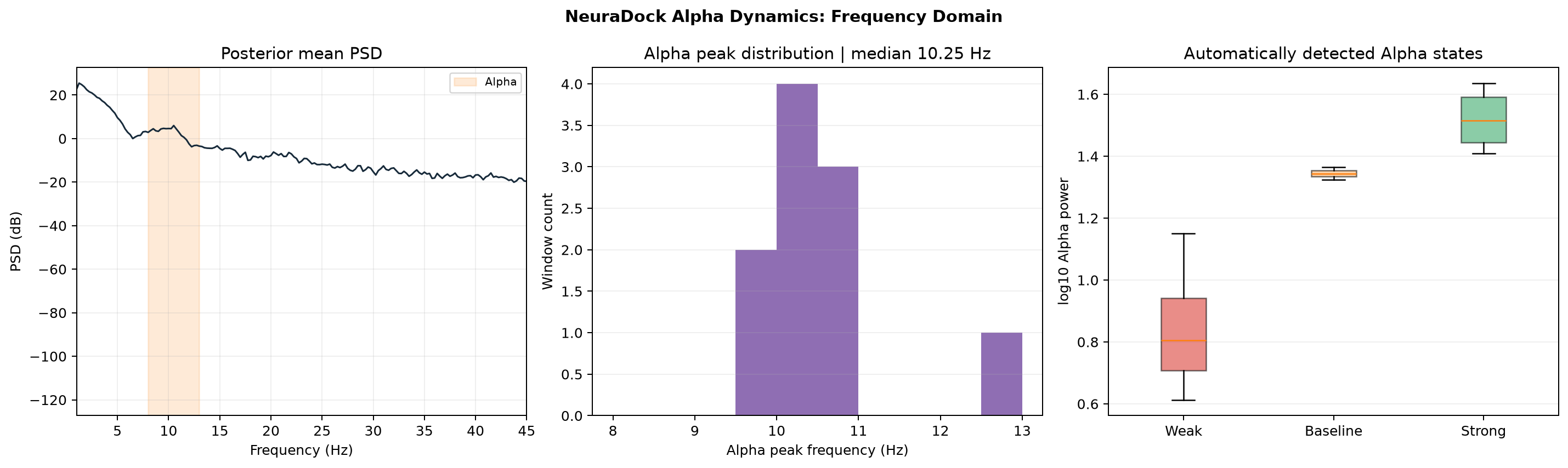}
  \caption{Time-domain and frequency-domain Alpha dynamics figures generated from the same \texttt{open\_closed\_eye2.txt} run.}
  \label{fig:alpha-time-frequency-supporting}
\end{figure}

\section{Step 4: Run Offline Rest/Task Visual Cognitive-Load Comparison}

For offline workload comparison, always compare a subject to their own baseline. Use the Rest file first and Task file second:

\begin{lstlisting}[style=cmd]
.\.venv\Scripts\neuradock-agent.exe analyze `
  data_examples\rest_task\rest_S01_1.txt `
  data_examples\rest_task\task_S01_1.txt `
  --workflow visual cognition comparison
\end{lstlisting}

With explicit labels:

\begin{lstlisting}[style=cmd]
.\.venv\Scripts\neuradock-agent.exe analyze `
  data_examples\rest_task\rest_S01_1.txt `
  data_examples\rest_task\task_S01_1.txt `
  --workflow visual cognition comparison `
  --condition-labels Rest Task
\end{lstlisting}

Typical output structure:

\begin{lstlisting}
runs/<timestamp>_visual_cognitive_load_comparison/
|-- report.md
|-- results.json
`-- figures/
    |-- visual_cognitive_load_comparison.png
    |-- rest_visual_cognitive_load.png
    `-- task_visual_cognitive_load.png
\end{lstlisting}

Interpretation rule:

\begin{lstlisting}
Task minus Rest posterior log Alpha < 0
  -> Task has lower posterior Alpha than Rest
  -> This is consistent with task-related Alpha suppression
\end{lstlisting}

This comparison is descriptive and within-subject. It should not be used to compare one subject's cognitive ability against another subject.

The actual tutorial run for \texttt{rest\_S01\_1.txt} versus \texttt{task\_S01\_1.txt} produced:

\begin{lstlisting}
Summary : visual_cognitive_load_comparison:
          Task-Rest median log Alpha = -0.024

Rest median posterior log Alpha : 1.141
Task median posterior log Alpha : 1.117
Task minus Rest                 : -0.024
Task / Rest Alpha power ratio   : 0.946
Rest median Alpha peak          : 11.00 Hz
Task median Alpha peak          : 11.25 Hz
Peak shift                      : +0.25 Hz
Rest retention                  : 93.3%
Task retention                  : 78.9%
Task quality                    : warning
\end{lstlisting}

The direction of the primary contrast is consistent with mild task-related posterior Alpha suppression: Task Alpha was lower than Rest Alpha. However, the effect is small and the Task condition had a quality warning, so the correct interpretation is cautious: this is a descriptive within-subject contrast, not proof of higher cognitive load.

\begin{figure}[htbp]
  \centering
  \includegraphics[width=\linewidth]{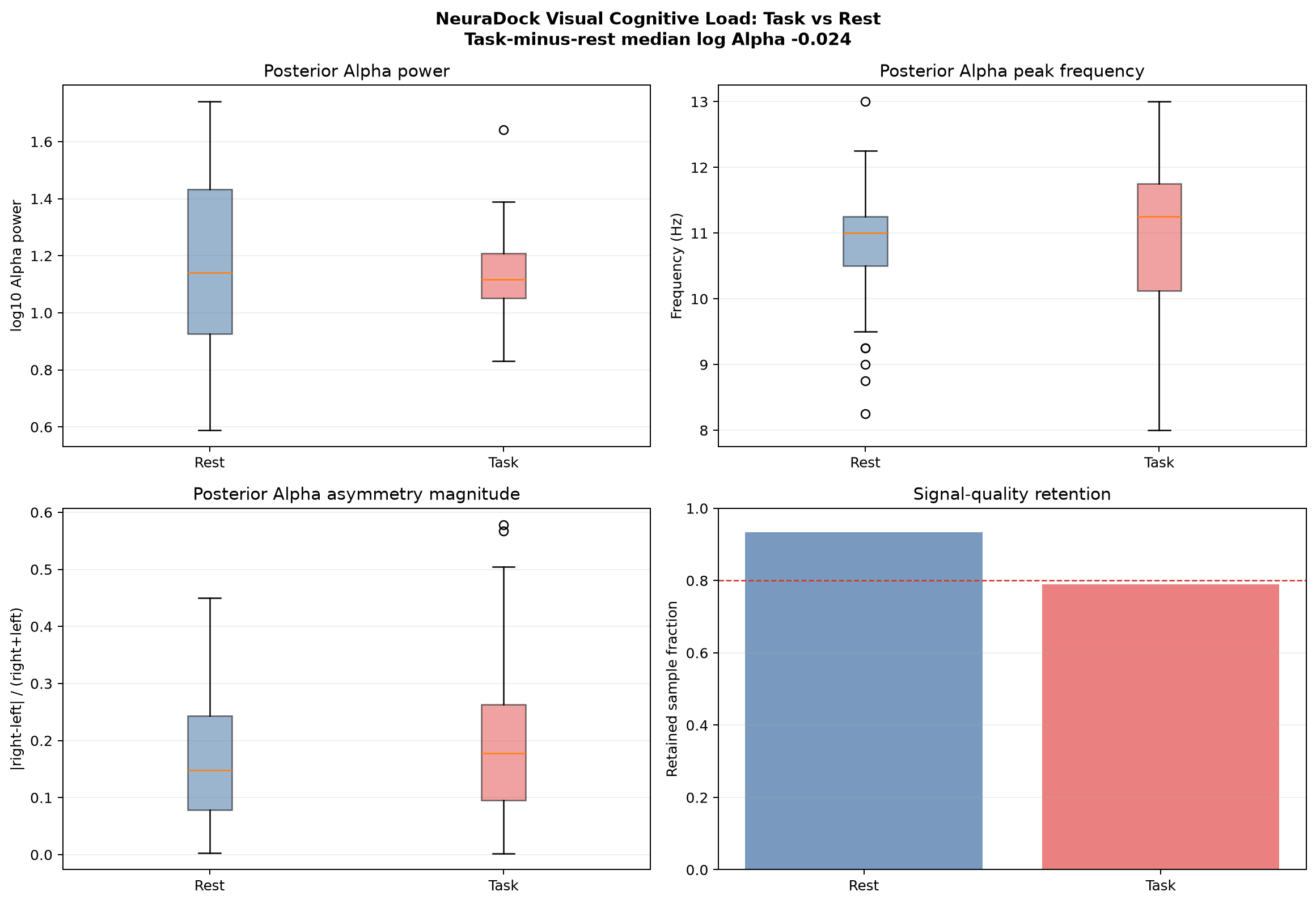}
  \caption{Actual Rest/Task comparison figure from \texttt{rest\_S01\_1.txt} and \texttt{task\_S01\_1.txt}. The primary contrast is the median posterior log Alpha difference between Task and Rest.}
  \label{fig:tutorial-rest-task-comparison}
\end{figure}

\begin{figure}[htbp]
  \centering
  \includegraphics[width=0.49\linewidth]{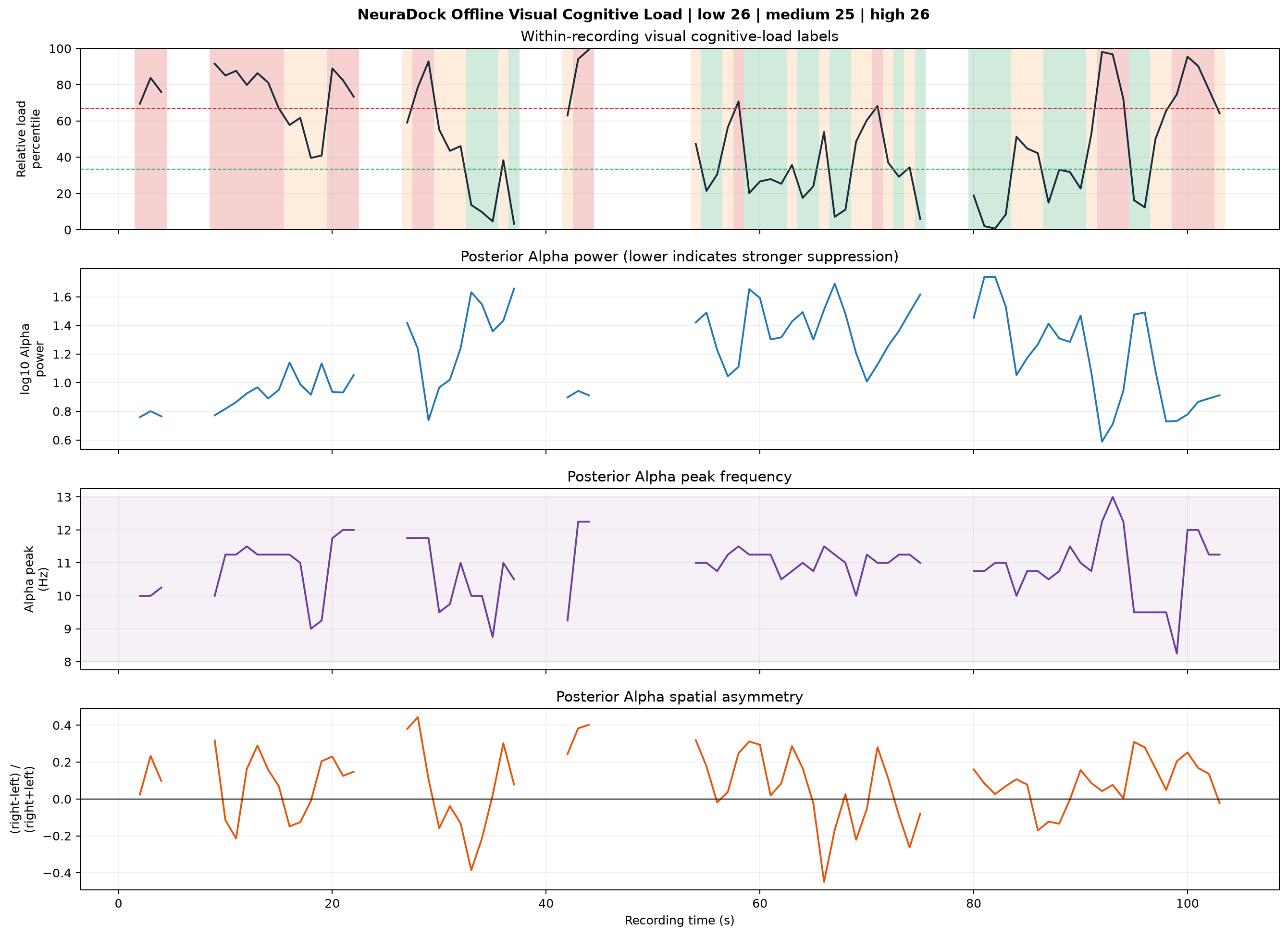}
  \includegraphics[width=0.49\linewidth]{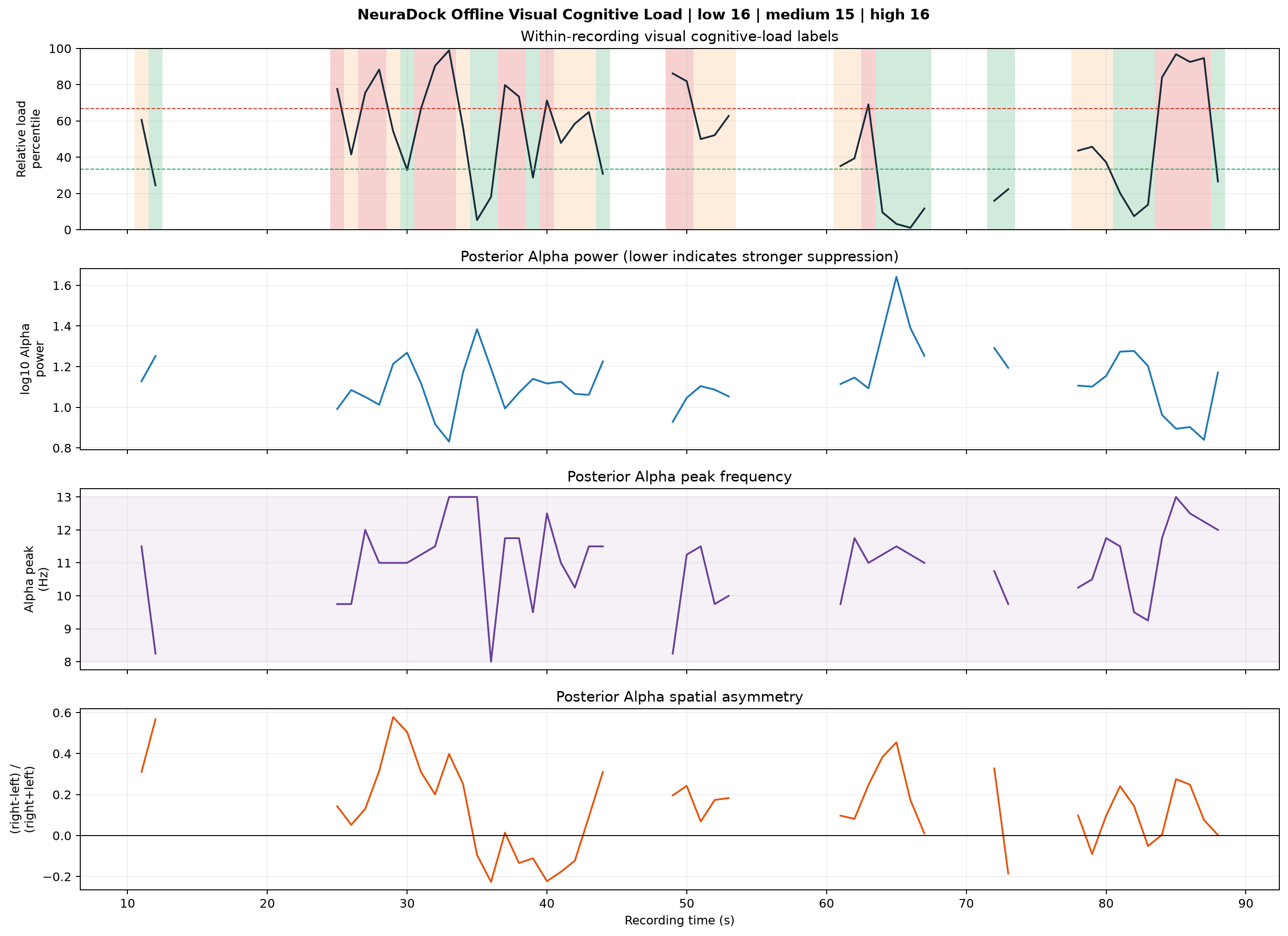}
  \caption{Condition-level visual cognitive-load plots for the Rest and Task recordings. Blank or disconnected regions correspond to QC-excluded windows, not zero workload.}
  \label{fig:tutorial-rest-task-condition-plots}
\end{figure}

\section{Step 5: Run the Public Mini-Dataset Analyses}

The complete human EEG mini-dataset is versioned in the separate public data
repository. Clone that branch next to the agent repository:

\begin{lstlisting}[style=cmd]
git clone --branch add-visual-cognitive-load-mini-dataset-20260622 `
  --single-branch `
  https://github.com/Neuradock/eeg-workstation-data.git `
  ..\eeg-workstation-data
\end{lstlisting}

Define a relative data root from the agent repository:

\begin{lstlisting}[style=cmd]
$dataRoot = "..\eeg-workstation-data\visual_cognitive_load\mini_dataset_v20260622"
\end{lstlisting}

Run Alpha Dynamics on all 18 recordings using the public CLI:

\begin{lstlisting}[style=cmd]
.\.venv\Scripts\neuradock-agent.exe analyze $dataRoot `
  --recursive `
  --workflow alpha dynamics `
  --output-root runs\mini_dataset_alpha
\end{lstlisting}

The following reusable PowerShell function runs a quality-gated,
within-subject comparison:

\begin{lstlisting}[style=cmd]
function Compare-NeuraDockPair {
  param($Reference, $Comparison, $ReferenceLabel, $ComparisonLabel)
  .\.venv\Scripts\neuradock-agent.exe analyze `
    $Reference $Comparison `
    --workflow visual cognition comparison `
    --condition-labels $ReferenceLabel $ComparisonLabel `
    --output-root runs\mini_dataset_comparisons
}
\end{lstlisting}

Run the six Rest/Task session comparisons:

\begin{lstlisting}[style=cmd]
foreach ($subject in "S01", "S02", "S03") {
  foreach ($session in 1, 2) {
    $folder = "$dataRoot\cohort_3subj_rest_task\$subject"
    Compare-NeuraDockPair `
      "$folder\rest_${subject}_${session}.txt" `
      "$folder\task_${subject}_${session}.txt" `
      "Rest" "Task"
  }
}
\end{lstlisting}

Run the four task-variant comparisons:

\begin{lstlisting}[style=cmd]
$cohort = "$dataRoot\cohort_2subj_ljw_xzy"
Compare-NeuraDockPair "$cohort\ljw\01_rest.txt" `
  "$cohort\ljw\02_chat.txt" "Rest" "Chat"
Compare-NeuraDockPair "$cohort\ljw\01_rest.txt" `
  "$cohort\ljw\03_game.txt" "Rest" "Game"
Compare-NeuraDockPair "$cohort\xzy\01_rest_eye_half.txt" `
  "$cohort\xzy\02_music_eye_half.txt" "Rest" "Music"
Compare-NeuraDockPair "$cohort\xzy\01_rest_eye_half.txt" `
  "$cohort\xzy\03_game.txt" "Rest" "Game"
\end{lstlisting}

Each command writes a timestamped run containing \texttt{report.md},
\texttt{results.json}, and quality-gated figures. The aggregate figures below
were assembled from those public per-recording and per-comparison outputs
during the reference validation. They are shown as a compact article summary;
the supported public CLI contract is the individual analysis output described
above.

The validation policy is intentionally conservative:

\begin{itemize}[leftmargin=*]
  \item No cross-subject comparisons are made.
  \item Each task is compared only to the same subject's own rest or baseline file.
  \item Mixed-eye recordings are retained as caveats rather than silently treated as clean protocol data.
  \item All reported Alpha and workload metrics are computed after preprocessing and QC gating.
\end{itemize}

\begin{figure}[htbp]
  \centering
  \includegraphics[width=\linewidth]{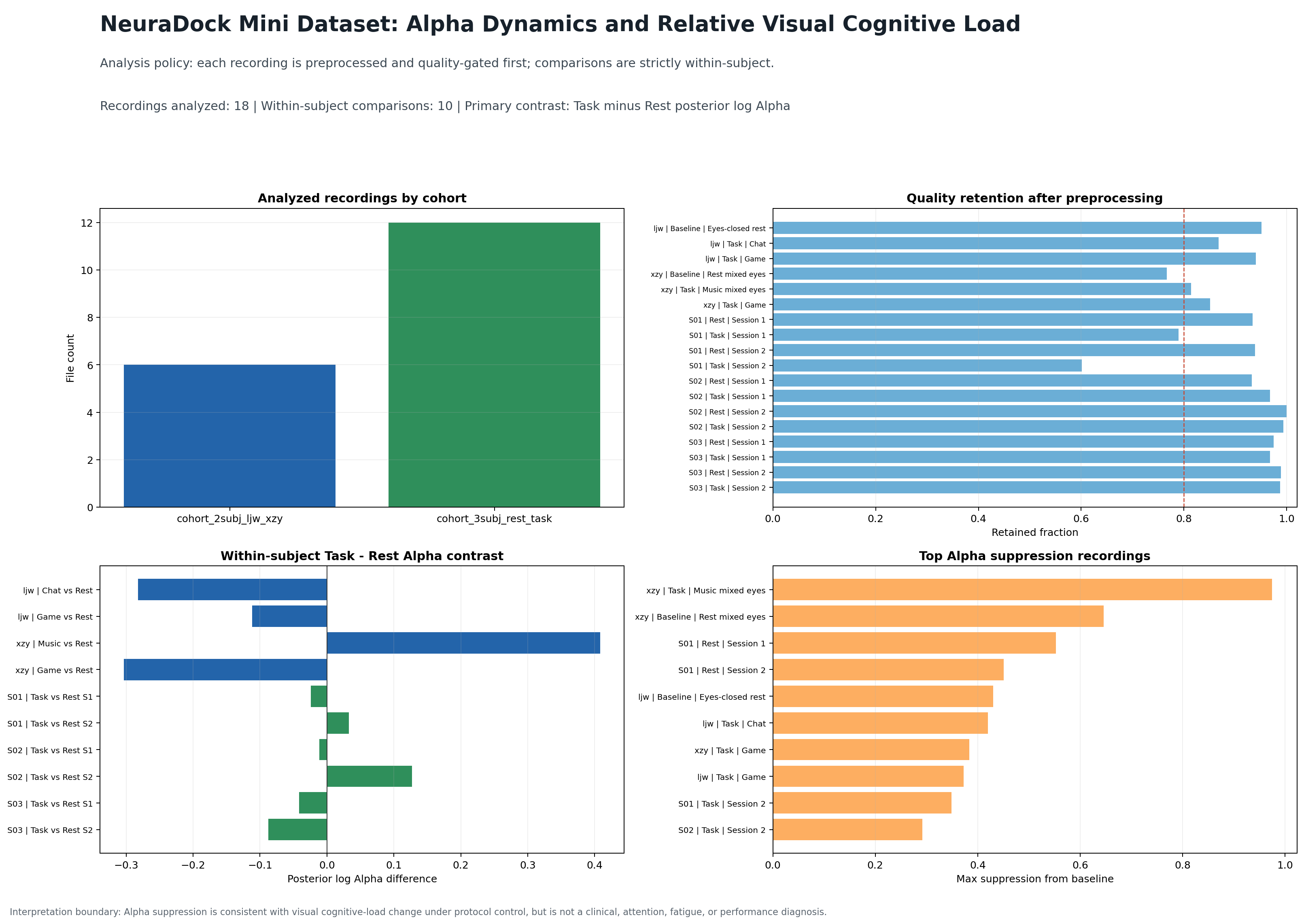}
  \caption{Reference mini-dataset summary assembled from the public quality-gated analysis outputs.}
  \label{fig:mini-professional-tutorial}
\end{figure}

\begin{figure}[htbp]
  \centering
  \includegraphics[width=\linewidth]{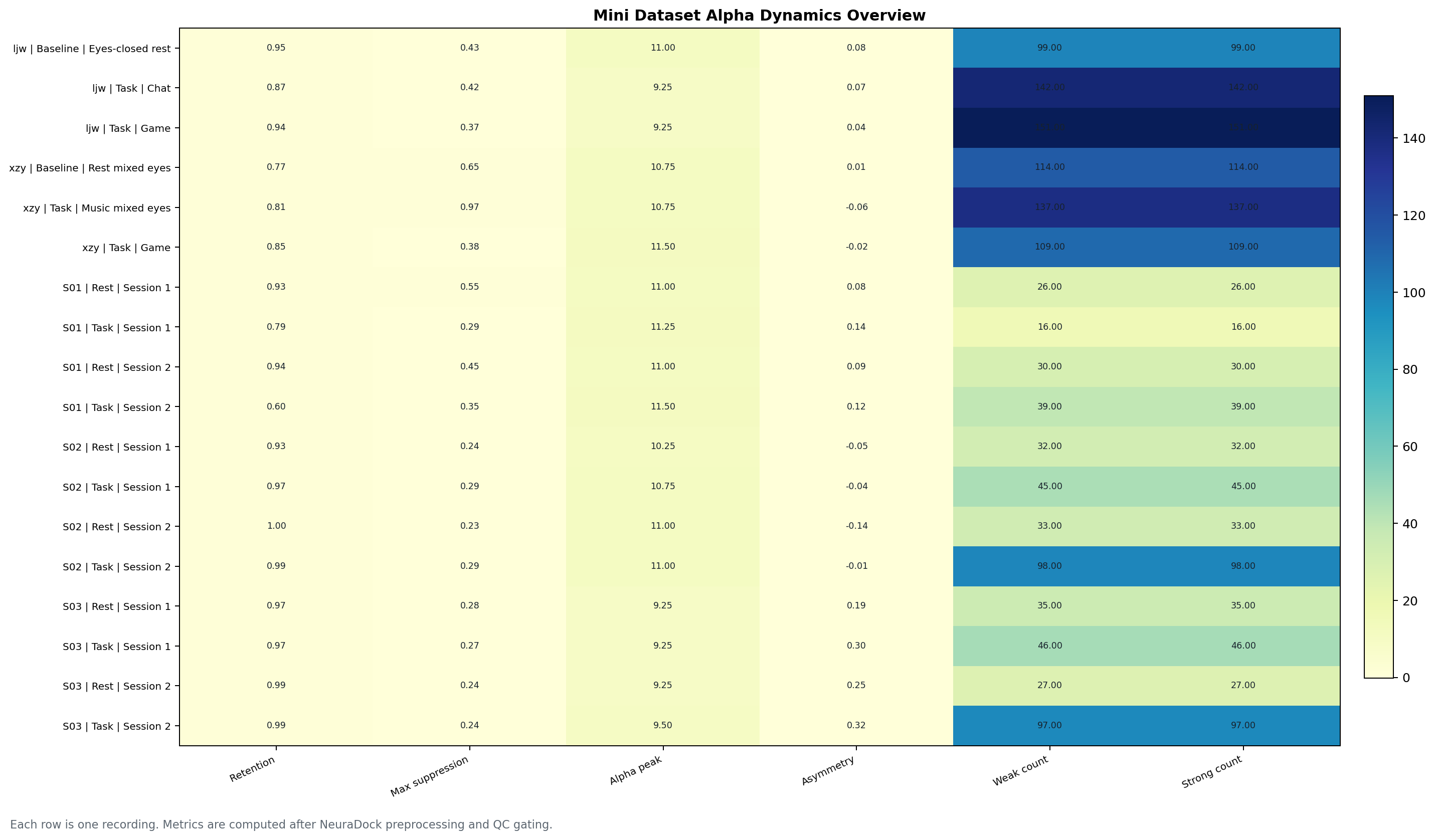}
  \caption{Reference Alpha dynamics overview across the mini-dataset. Each row is one recording after preprocessing and QC gating.}
  \label{fig:mini-alpha-overview-tutorial}
\end{figure}

\begin{figure}[htbp]
  \centering
  \includegraphics[width=\linewidth]{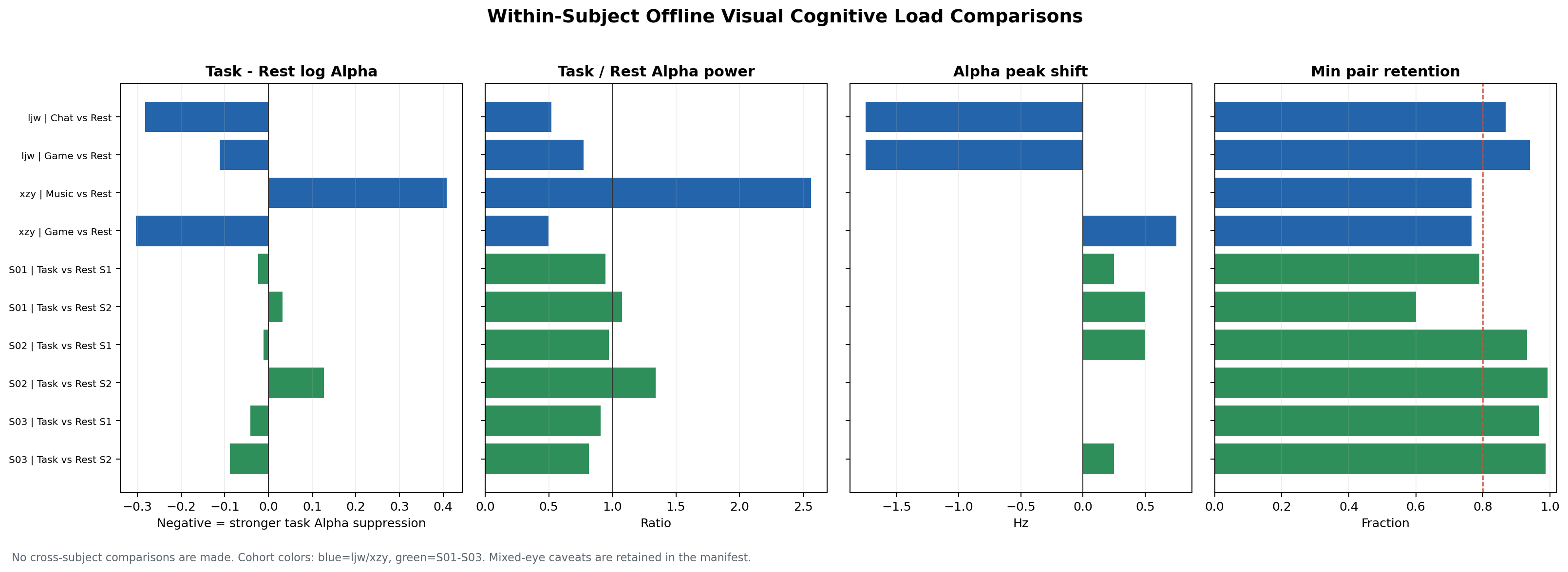}
  \caption{Reference within-subject comparison overview. Negative task-minus-rest log Alpha indicates stronger task-related Alpha suppression.}
  \label{fig:mini-comparison-overview-tutorial}
\end{figure}

\subsection{Mini-Dataset Results to Check}

The reference run analyzed 18 recordings and generated 10 within-subject comparisons. Seven of 10 contrasts showed lower task posterior Alpha than rest. In the most interpretable cases, task-related posterior Alpha suppression reached approximately 50\% power reduction, consistent with the expected direction from visual-load Alpha literature. The clearest examples were:

\begin{itemize}[leftmargin=*]
  \item \texttt{xzy Game vs Rest}: Task minus Rest log Alpha = -0.303, Task/Rest Alpha ratio = 0.50.
  \item \texttt{ljw Chat vs Rest}: Task minus Rest log Alpha = -0.283, Task/Rest Alpha ratio = 0.52.
  \item \texttt{ljw Game vs Rest}: Task minus Rest log Alpha = -0.112, Task/Rest Alpha ratio = 0.77.
\end{itemize}

The \texttt{xzy Music vs Rest} contrast moved in the opposite direction and should be interpreted cautiously because of the mixed-eye caveat. This is a useful tutorial outcome: a credible tool should reveal risk cases and unexpected directions rather than force every task into a high-load result.

Baseline retest analysis for \texttt{S01--S03} Rest session 1 versus Rest session 2 showed Pearson \(r = 0.803\) and ICC(C,1) = 0.765 for median posterior log Alpha. This is initial evidence of within-subject repeatability, not a population-level reliability claim.

\begin{table}[htbp]
\centering
\caption{How to interpret the mini-dataset validation.}
\label{tab:what-this-proves}
\small
\begin{tabular}{p{0.46\linewidth}p{0.46\linewidth}}
\toprule
What this proves & What this does not prove \\
\midrule
The agent can run an end-to-end, QC-gated workflow on multiple local EEG files and produce reproducible Alpha and workload figures. & The mini-dataset does not establish population norms, clinical validity, or universal thresholds for every user and task. \\
Within-subject Rest/Task contrasts can reveal posterior Alpha suppression in interpretable visual-task cases. & A lower Alpha value alone does not prove cognitive load without protocol context, behavior, and quality review. \\
Baseline retest results provide initial evidence of within-subject repeatability for the same subject and condition. & The retest result is not a full reliability study across demographics, hardware sessions, and task designs. \\
\bottomrule
\end{tabular}
\end{table}

\section{Step 6: Start the Online Dashboard}

For real NeuraDock hardware, the user only needs the hardware stream IP and port:

\begin{lstlisting}[style=cmd]
.\.venv\Scripts\neuradock-agent.exe online `
  --ip 192.168.4.1 `
  --port 9600
\end{lstlisting}

The agent will connect to the stream, send the device start command, parse packets, run online preprocessing and QC, calculate rolling visual workload, and open the local dashboard:

\begin{lstlisting}
http://127.0.0.1:8765
\end{lstlisting}

For a local demo without hardware, run:

\begin{lstlisting}[style=cmd]
.\.venv\Scripts\neuradock-agent.exe serve --port 8765
\end{lstlisting}

When no \texttt{--demo-file} is supplied, this command generates a
deterministic synthetic replay. A reader can therefore test the UI without
NeuraDock hardware or human EEG data. Open:

\begin{lstlisting}
http://127.0.0.1:8765
\end{lstlisting}

If the dashboard is open but no points are visible yet, press \texttt{Start} in the UI or advance the demo stream manually:

\begin{lstlisting}[style=cmd]
Invoke-RestMethod http://127.0.0.1:8765/api/demo/next
\end{lstlisting}

The tutorial screenshot below was captured from the local demo dashboard. It shows a rolling load index, Alpha state, Alpha peak, quality status, Alpha metrics, and the online preprocessing panel.

\begin{figure}[htbp]
  \centering
  \includegraphics[width=\linewidth]{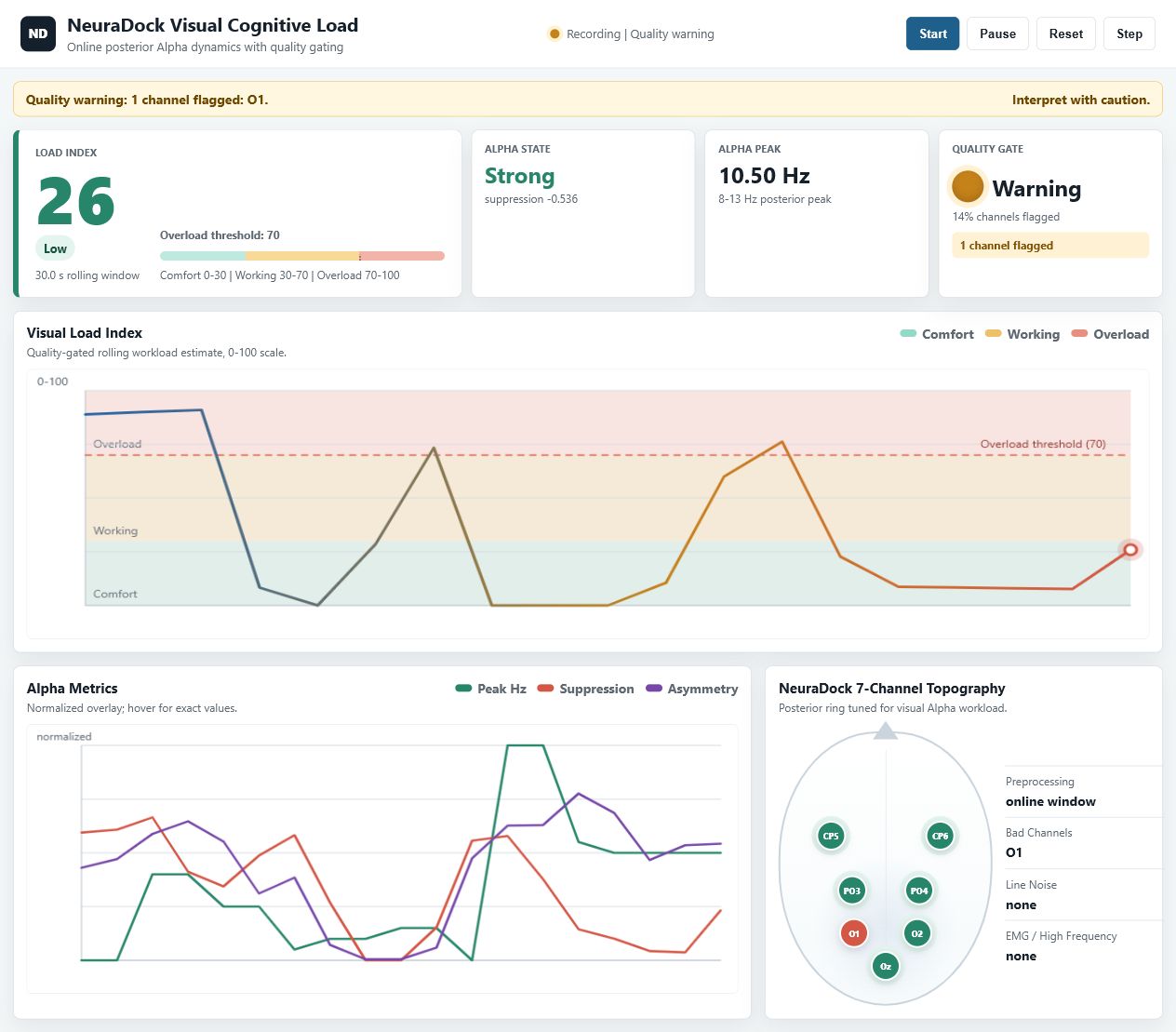}
  \caption{Local demo dashboard UI. This screenshot does not require hardware; the current public command uses a deterministic synthetic replay.}
  \label{fig:tutorial-dashboard-ui}
\end{figure}

Dashboard fields to inspect:

\begin{itemize}[leftmargin=*]
  \item Rolling visual workload index.
  \item Alpha state: weak Alpha, baseline Alpha, or strong Alpha.
  \item Alpha peak frequency.
  \item Alpha suppression from rolling baseline.
  \item Quality status and bad-channel candidates.
  \item Stream status and samples received.
\end{itemize}

The online parser is tuned for the NeuraDock stream profile: 250 Hz sampling, 5 samples per Bluetooth packet, a TCP bridge, and a 4 s rolling analysis window. Another EEG system can be connected only if its stream is adapted into the expected seven-channel sample matrix and packet semantics. Otherwise, the parser and QC assumptions should be treated as hardware-specific implementation details rather than generic EEG defaults.

\section{Step 7: Use the Real-Time API in an Application}

After starting \texttt{online} or \texttt{serve}, check status:

\begin{lstlisting}[style=cmd]
Invoke-RestMethod http://127.0.0.1:8765/api/status
\end{lstlisting}

User applications can read the latest computed workload from:

\begin{lstlisting}
GET http://127.0.0.1:8765/api/status
\end{lstlisting}

Important fields:

\begin{lstlisting}
status
current.visual_load_index
current.alpha_state
current.alpha_peak_hz
current.alpha_suppression_from_baseline
current.alpha_asymmetry_right_minus_left
current.quality_status
quality.status
quality.bad_channel_candidates
stream.connected
stream.samples_received
\end{lstlisting}

Minimal Python integration:

\begin{lstlisting}[language=Python]
import requests

data = requests.get("http://127.0.0.1:8765/api/status", timeout=2).json()
if data["status"] == "ok" and data["quality"]["status"] == "pass":
    workload = data["current"]["visual_load_index"]
    alpha_state = data["current"]["alpha_state"]
    alpha_peak = data["current"]["alpha_peak_hz"]
    print(workload, alpha_state, alpha_peak)
\end{lstlisting}

Recommended integration pattern:

\begin{lstlisting}
NeuraDock hardware
  -> neuradock-agent online
  -> GET /api/status
  -> user application
\end{lstlisting}

\subsection{Latency Benchmark}

The reference tutorial run used a local replay with a 4 s rolling window and 1 s step. Quality-valid online steps showed median core processing latency of 1.89 ms and p95 latency of 2.66 ms. The local HTTP demo endpoint showed median latency of 15.15 ms and p95 latency of 27.18 ms. The current no-hardware command uses a deterministic synthetic replay, so these historical measurements should be treated as reference values rather than guaranteed deployment latency.

This benchmark measures local computation and API serving, not physical wireless latency, browser rendering, or external application delay. The practical update cadence is dominated by the configured analysis step and window length.

\subsection{Industrial Translation}

For teams building adaptive UI, XR interaction prototypes, rehabilitation-device workflows, driver-monitoring or HMI research systems, and learning dashboards, the useful interface is deliberately simple. The \texttt{/api/status} endpoint returns a 0--100 \texttt{current.visual\_load\_index} and a quality flag at the online update cadence. Applications should gate interventions on \texttt{quality.status} and \texttt{current.quality\_status} being \texttt{pass} before acting on the load estimate.

A typical integration rule is:

\begin{lstlisting}
if quality.status == "pass"
   and current.quality_status == "pass"
   and current.visual_load_index > threshold
   for N consecutive windows:
       adapt interface
else:
       hold state, continue monitoring, or request recalibration
\end{lstlisting}

This is the commercial translation of the agent's design: the same QC gate used for research figures can also prevent an application from reacting to noisy real-time estimates.

\section{Step 8: Ask for LLM Interpretation}

The LLM mode is an explanation layer. It does not compute EEG features and does not receive raw EEG arrays. It receives allowlisted summaries, warnings, and interpretation limits.

Example offline explanation command:

\begin{lstlisting}[style=cmd]
.\.venv\Scripts\neuradock-agent.exe ask `
  "Compare Rest and Task visual cognitive load and explain quality risks" `
  --file data_examples\rest_task\rest_S01_1.txt `
  --file2 data_examples\rest_task\task_S01_1.txt `
  --llm `
  --output-root runs\tutorial_llm
\end{lstlisting}

Example Alpha dynamics explanation:

\begin{lstlisting}[style=cmd]
.\.venv\Scripts\neuradock-agent.exe ask `
  "Analyze Alpha dynamics and explain cognitive-load risk limits" `
  --file data_examples\alpha\open_closed_eye2.txt `
  --llm
\end{lstlisting}

A good LLM interpretation should say what the computed values suggest, what quality risks exist, and what conclusions should not be drawn. For example, if \texttt{quality.status} is warning or a posterior channel is listed as a bad-channel candidate, the summary should reduce confidence rather than hide the issue.

\subsection{Reference LLM Interpretation}

One reference tutorial run used an OpenAI-compatible API endpoint and model
\texttt{qwen3.7-max}. A current run writes the interpretation under the
selected output root:

\begin{lstlisting}
runs/tutorial_llm/
  <run-id>_visual_cognitive_load_comparison/
    llm_interpretation.md
\end{lstlisting}

The archived June 2026 reference file reported the metadata below. Its context
version predates the current \texttt{2026.6.24} release:

\begin{lstlisting}
Status: success
Model: qwen3.7-max
Prompt version: neuradock-result-interpretation-v3
Context version: 2026.6.18
Generated at: 2026-06-22T11:18:12+08:00
Raw EEG sent to model: no
\end{lstlisting}

The following excerpt is copied from the real generated LLM interpretation, with the warning symbol removed for LaTeX portability:

\begin{quote}
\textbf{Data Quality Warning.} Signal quality is limited in both conditions, particularly during the Task. The Task recording retained only 78.9\% of samples after segment QC, and the system detected slight body activity or mild muscle tension. Forty out of 87 Task windows were excluded because they did not meet the 80\% clean-sample requirement. The Rest recording had better sample retention at 93.3\%, but 25 out of 102 windows were still excluded. Because nearly half of the Task windows and a quarter of the Rest windows were excluded, temporal coverage is sparse. This reduces confidence in continuous tracking of cognitive load.

\textbf{Main Findings.} The median posterior log Alpha power was slightly lower during Task (1.12) compared with Rest (1.14), yielding a Task-to-Rest power ratio of 0.95 and a difference of -0.024. This small decrease is consistent with Alpha suppression, a common sensor-level pattern when transitioning from a relaxed state to a visual or cognitive task. The median Alpha peak frequency shifted slightly higher during Task, from 11.0 Hz to 11.25 Hz. Posterior left/right Alpha asymmetry magnitude was also slightly higher during Task.

\textbf{Boundary.} Lower Task posterior Alpha is consistent with stronger Alpha suppression, but it does not by itself prove higher cognitive load or establish a causal conclusion. Visual cognitive-load labels and indices are relative, within-recording research heuristics. They are not medical, psychological, attention, fatigue, or performance diagnoses.
\end{quote}

This real output illustrates the intended LLM role: it summarizes deterministic EEG results, foregrounds quality risks, and states scientific boundaries. It does not compute the Alpha features and it does not receive raw EEG.

\section{Troubleshooting}

\subsection{PowerShell Prints a profile.ps1 Warning}

Some Windows environments print a warning such as:

\begin{lstlisting}
profile.ps1 cannot be loaded because running scripts is disabled
\end{lstlisting}

This warning comes from the local PowerShell profile loading policy. It does not necessarily mean the NeuraDock command failed. Check the actual command output and generated files.

\subsection{LLM Output File Exists but PowerShell Shows an Encoding Error}

On some Windows systems, PowerShell may fail to print Unicode symbols from an LLM interpretation, for example:

\begin{lstlisting}
'gbk' codec can't encode character ...
\end{lstlisting}

If \texttt{llm\_interpretation.md} was created, the LLM call itself succeeded. Open the Markdown file in a UTF-8 editor, or read it with:

\begin{lstlisting}[style=cmd]
Get-Content runs\YOUR_RUN_ID\llm_interpretation.md -Encoding UTF8
\end{lstlisting}

The tutorial run encountered this exact console-encoding issue after successfully saving the LLM interpretation file.

\subsection{The Dashboard Port Is Already Used}

Use a different dashboard port:

\begin{lstlisting}[style=cmd]
.\.venv\Scripts\neuradock-agent.exe online `
  --ip 192.168.4.1 `
  --port 9600 `
  --dashboard-port 8766
\end{lstlisting}

\subsection{The Online API Has No Data}

Check:

\begin{itemize}[leftmargin=*]
  \item Whether the hardware IP and port are correct.
  \item Whether the stream is connected in \texttt{/api/status}.
  \item Whether samples are being received.
  \item Whether the buffer is still warming up for the first 4 s window.
  \item Whether QC warnings are preventing high-confidence interpretation.
\end{itemize}

\subsection{Retention Is Low}

Low retention means a large fraction of samples or windows failed QC. Treat downstream Alpha or workload estimates as lower confidence. Inspect \texttt{report.md}, \texttt{results.json}, and quality figures before interpreting the result.

\subsection{Mixed-Eye Data Needs Caution}

Eye state strongly affects posterior Alpha. Mixed-eye recordings can be useful for stress-testing the agent, but they should not be interpreted like clean eyes-open or eyes-closed protocol data. The mini-dataset intentionally keeps mixed-eye caveats visible.

\subsection{Running Without Hardware}

Readers without NeuraDock hardware can still run most of the tutorial.
Synthetic replay supports the demo dashboard, \texttt{/api/status}, and
\texttt{/api/demo/next}; downloaded public TXT files support preprocessing, QC
reports, Alpha dynamics, offline Rest/Task comparison, mini-dataset analysis,
and LLM interpretation. These modes cannot validate the physical wireless
path: Bluetooth throughput, TCP bridge behavior, packet loss, device start
command behavior, or end-to-end hardware latency.

\subsection{Dependency Conflicts}

For a clean developer environment, use a fresh virtual environment:

\begin{lstlisting}[style=cmd]
Remove-Item -Recurse -Force .venv
py -m venv .venv
.\.venv\Scripts\python.exe -m pip install --upgrade pip
.\.venv\Scripts\python.exe -m pip install -e ".[dev]"
\end{lstlisting}

Then run the focused tests:

\begin{lstlisting}[style=cmd]
.\.venv\Scripts\python.exe -m pytest -q `
  tests\test_workflows.py tests\test_online.py
\end{lstlisting}

\subsection{Docker Status}

This first open-source release does not bundle an official Docker image. The supported reproducible path is the Python virtual environment shown in Step 1. A future Docker workflow should preserve the same CLI entry points and keep raw EEG local unless the user explicitly configures external services.

\subsection{Extending or Contributing}

Developers who want to extend the workflow should start from \texttt{COMMANDS.md}, \texttt{CONTRIBUTING.md}, \texttt{src/neuradock\_agent/workflows.py}, and \texttt{src/neuradock\_agent/cli.py}. New analysis features should keep the same rule as the built-in workflows: compute downstream metrics from the preprocessing/QC-gated path and expose quality risks alongside any workload estimate.

\subsection{pdflatex Is Not Installed}

The tutorial source is a standard \LaTeX{} file. To render a PDF, install a TeX distribution such as TeX Live or MiKTeX, open a terminal in the directory containing this file, then run:

\begin{lstlisting}[style=cmd]
pdflatex neuradock_visual_cognitive_load_agent_tutorial_20260618.tex
pdflatex neuradock_visual_cognitive_load_agent_tutorial_20260618.tex
\end{lstlisting}

\section{Discussion}

\subsection{What the Tutorial Demonstrates}

This tutorial demonstrates an offline-to-online bridge. A reader can start from a static EEG file, pass through preprocessing and QC, inspect Alpha dynamics, compare rest and task within subject, run the public mini-dataset analyses and compare them with the reference summary, and then use the same conceptual pipeline in a real-time dashboard and API.

The most important scientific design choice is that all downstream interpretation depends on quality-gated data. This makes the agent more conservative but more credible. The agent does not simply output a workload score; it reports whether the signal quality supports interpretation.

The hardware design is part of the same argument. The posterior NeuraDock montage concentrates sensors around visual Alpha generators, while the online parser and rolling-window defaults match the NeuraDock data-throughput profile. This makes the tutorial more than a generic EEG script collection: it is a hardware-tuned, application-facing workflow for visual cognitive-load prototypes.

\subsection{Comparison With Existing Tools}

\begin{table}[htbp]
\centering
\caption{Tutorial positioning relative to existing EEG tools and device ecosystems.}
\label{tab:tutorial-tool-comparison}
\small
\resizebox{\linewidth}{!}{%
\begin{tabular}{lllll}
\toprule
Tool & General vs focused & Offline vs real-time & Open source & Cognitive-load specific \\
\midrule
MNE-Python & General & Primarily offline & Yes & No \\
BrainFlow & General & Real-time acquisition & Yes & No \\
EEGLAB & General & Primarily offline & Yes & No \\
Emotiv SDK/Cortex & General ecosystem & Real-time API & No/vendor & No \\
Neurosity SDK & General ecosystem & Real-time API & SDK-oriented & No \\
NeuraDock Agent & Focused & Offline and real-time & Yes & Yes \\
\bottomrule
\end{tabular}
}
\end{table}

This table is not a replacement claim. MNE-Python, BrainFlow, and EEGLAB remain valuable foundations. NeuraDock Agent is useful because it narrows the workflow to a specific applied problem: visual cognitive load from quality-gated Alpha dynamics.

\subsection{Limitations}

The mini-dataset is small and should be understood as a tutorial validation set, not a population-level scientific study. The online benchmark used local demo replay rather than physical wireless hardware. The current workload estimate is primarily based on posterior Alpha dynamics and does not yet integrate behavioral accuracy, reaction time, subjective workload scales, pupil size, or multimodal physiology. Larger protocol-controlled validation is necessary before deployment in high-stakes contexts.

\section{Conclusion}

NeuraDock Agent is best understood as a focused, open-source tutorial workflow for visual cognitive-load EEG. It guides a reader from installation to preprocessing, Alpha dynamics, offline within-subject comparison, public mini-dataset analysis, online dashboard/API use, and LLM explanation. The tutorial format makes the system's value concrete: readers can run the commands, inspect the generated figures, verify that metrics are quality-gated, and integrate the real-time API into their own applications. For developers extending the workflow, \texttt{COMMANDS.md} and \texttt{CONTRIBUTING.md} provide the starting point; new features should preserve the agent's central promise: cognitive-load estimates are useful only when paired with transparent signal-quality evidence.

\section*{Reproducibility Checklist}

\begin{itemize}[leftmargin=*]
  \item Open-source repository:
    \url{https://github.com/Neuradock/eeg-workstation-agent}
  \item Public data repository:
    \url{https://github.com/Neuradock/eeg-workstation-data}
  \item Main command guide: \texttt{COMMANDS.md}
  \item Contributor guide: \texttt{CONTRIBUTING.md}
  \item Data usage rules: \texttt{DATASET.md}
  \item Verified example downloader:
    \texttt{scripts/download\_example\_data.py}
  \item Workflow implementation: \texttt{src/neuradock\_agent/workflows.py}
  \item CLI routing: \texttt{src/neuradock\_agent/cli.py}
  \item Realtime API implementation:
    \texttt{src/neuradock\_agent/online.py}
  \item Automated tests: \texttt{tests/}
  \item This tutorial source and its adjacent \texttt{figures/} directory.
\end{itemize}

\section*{Intended Use Statement}

NeuraDock Agent outputs are relative research and engineering signals. They are not medical, clinical, attention, fatigue, cognitive-ability, or performance diagnoses. Any deployment that affects users should include protocol control, quality review, informed consent, privacy safeguards, and independent validation.

\end{document}